\documentclass{article}

\usepackage{arxiv}

\usepackage[utf8]{inputenc} % allow utf-8 input
\usepackage[T1]{fontenc}    % use 8-bit T1 fonts
\usepackage{hyperref}       % hyperlinks
\usepackage{url}            % simple URL typesetting
\usepackage{booktabs}       % professional-quality tables
\usepackage{amsfonts}       % blackboard math symbols
\usepackage{nicefrac}       % compact symbols for 1/2, etc.
\usepackage{microtype}      % microtypography
\usepackage{lipsum}

\usepackage{amssymb}
\usepackage{amsmath}
\usepackage{bbm}
\usepackage{dsfont}
\usepackage{mathrsfs}

\usepackage{graphicx}
\usepackage{subcaption}
\usepackage{bm}

\newcommand{\rw}{\mathrm{w}}
\newcommand{\vx}{\mathbf{x}}

\newcommand{\nmathbf}{\bm}

\newcommand{\sD}{\mathscr{D}_n}

\newcommand{\sH}{\mathscr{H}}

\newcommand{\sX}{\mathscr{X}}

\newcommand{\one}{\mathbbm{1}}

\newcommand{\Real}{\mathbb{R}}
\newcommand{\krnl}{\mathcal{K}}

\def\bfw{\nmathbf w}
\def\bfA{\nmathbf A}

\def\bfD{\nmathbf D}
\def\bfE{\nmathbf E}
\def\bfG{\nmathbf G}
\def\bfI{\nmathbf I}
\def\bfK{\nmathbf K}
\def\bfL{\nmathbf L}
\def\bfM{\nmathbf M}
\def\bfN{\nmathbf N}

\def\bfY{\nmathbf Y}
\def\bfV{\nmathbf V}

\def\bfH{\nmathbf H}

\def\bfalpha{\nmathbf \alpha}

\title{Graph Enhanced High Dimensional Kernel Regression}

\author{
  Eddie Pei \\
  School of Mathematical Sciences\\
  Rochester Institute of Technology\\
  85 Lomb Memorial Drive, Rochester, New York 14623 \\
  \texttt{ep2667@rit.edu} \\
   \And
 Ernest Fokou\'e \\
  School of Mathematical Sciences\\
  Rochester Institute of Technology\\
  85 Lomb Memorial Drive, Rochester, New York 14623 \\
  \texttt{epfeqa@rit.edu} \\
}

\begin{document}
\maketitle

\begin{abstract}
  In this paper, the flexibility, versatility and predictive power of kernel regression are combined with now lavishly available network data to create regression models with even greater predictive performances. Building from previous work featuring generalized linear models built in the presence of network cohesion data, we construct a kernelized extension that captures subtler nonlinearities in extremely high dimensional spaces and also  produces far better predictive performances. Applications of seamless yet substantial adaptation to simulated and real-life data demonstrate the appeal and strength of our work. 
\end{abstract}

% keywords can be removed

\section{Introduction}
Graph theoretic methods have come under intense scrutiny in recent years thanks to the availability of ever increasing volumes of network data from all walks of life.  The literature on the theory, methodology, computation and applications involving graph theoretic methods now abounds. \cite{Newman:network:1}, \cite{Marques:graph:2017}, \cite{Shen:2017:Graph}, \cite{Meghanathan:2019}, \cite{kernel:network}. Regression analysis being one of the central pillars of supervised statistical machine learning, it is no surprise at all that many authors have already proposed and explored the incorporation of network data into regression learning with the finality of both interpretability and accurate prediction.  \cite{li2019} whose work rightly deserves to be deemed seminal in this context, provide a detail account of linear regression with network cohesion, specifically deriving both the algorithmic details both also a sound theory and a promising application. \cite{Li:randnet:2019} even provides a package in the {\sf R} software environment to help with the exploration of a variety of network data related statistical machine learning activities. Building from \cite{li2019} whose fundamental building block in the linear regression model later extended to the generalized linear models for a wider applicability, this present paper ignores the interpretability motivating \cite{li2019}'s adoption of the GLM family of models, and instead focuses on the versatility and the flexibility and the predictive power of the function space by choosing kernel regression as the fulcrum.  More specifically, in a spirit similar to \cite{li2019}, we build a regularized kernel regression framework with a regularizing component induced by a graph theoretic constraints assumed to underlie the accompanying network data. Essentially this paper's quintessential contribution lies in boosting the predictive performance of kernel regression by way of appropriately induced graph theoretic constraints. Indeed given regression data $\sD= \{(\vx_1,y_1) \overset{iid}{\sim} p_{xy}(\vx,y),\,\, \vx_i \in \sX,\, y_i\in \Real, \,\, i=1,\cdots,n\}$, we shall herein consider regression learning models under the homoscedastic Gaussian noise assumption, more specifically
\begin{eqnarray}
\label{eq:basic:gaussian:regr:1}
p(y_i | f, \vx_i, \sigma^2) = \phi(y_i; f(\vx_i), \sigma^2) = \frac{1}{\sqrt{2\pi\sigma^2}}e^{-\frac{1}{2}\frac{(y_i-f(\vx_i)^2}{\sigma^2}}
\end{eqnarray}
where $f \in \sH_{\krnl}$ with the function space $\sH_{\krnl}$ given here by 
\begin{eqnarray}
\label{eq:kernel:space:1}
\sH_{\krnl}=\Big\{f: \, \forall \vx \in \sX, f(\vx)=\alpha + \sum_{j=1}^{n}{\rw_j \krnl(\vx, \vx_j)},\,\, \alpha\in \Real,\, \rw_j\in \Real,\, j=1,\cdots,n\Big\}.
\end{eqnarray}

Our choice of the function space \eqref{eq:kernel:space:1} is motivated by our intention and desire to extend and improve \cite{li2019} both predictively and in terms of modelling flexibility and versatility. Indeed, the use of kernel regression family of models affords greater flexibility and versatility in modelling since one can seamlessly account and deal with models built on non real-valued input spaces. The added benefit being also the ability of kernel regression learning machines to represent arbitrarily complex underlying regression functions. In fact, kernel regression and kernel methods in general have grown tremendously in popularity, and are continually used and further developed and improved all the time. Kernel methods have never ceased to gain more popularity in modelling, and many software packages are continually built around them  \cite{kernel:network}. \cite{kernel:package:r:1}. Gaussian Processes for instance have provided one of the most commonly used kernel methods, and have become fashionable in machine learning, having proven to be formidable learning machines with excellent predictive performances on high dimensional \cite{GaussProc:2000}, \cite{Fokoue:2008},\cite{Fokoue:Goel:2006}, \cite{Fokoue:2009}, \cite{quinonero2005unifying}, \cite{GP:Book:1}. The phenomenal success of support vector classifiers was immediately followed by the arrival on the scene of support vector regression, with the seminal paper featuring support vector regression appearing in \cite{NIPS1996_1238} and \cite{Vapnik:SVM:1}. The relevance vector machine \cite{Tipping:2001} is yet another example of a kernel learning machine that yielded excellent predictive performances in high dimensional regression. Central to all the kernel learning machines is the need for suitable regularization owing to their inherent illposedness. It is found that in their practical form used for actually building the kernel regression machine, the estimation/learning task is done via 
\begin{eqnarray}
\label{eq:regularized:HK:1}
\widehat{R}_{\lambda,n}(f) = \frac{1}{2n}\sum_{i=1}^n{(y_i-f(\vx_i)^2} + \frac{\lambda}{2}\|f\|_{\krnl}^2
\end{eqnarray}

For the function space given in \eqref{eq:kernel:space:1}, the representer theorem of \cite{Wahba:1990}, \cite{Wahba:1998} leads to 
\begin{eqnarray}
\label{eq:regularized:HK:2}
\widehat{R}_{\lambda,n}(\bfw) = \frac{1}{2n}(\bfY-\alpha\one_n - \bfK\bfw)^\top(\bfY-\alpha\one_n - \bfK\bfw) + \frac{\lambda}{2}\bfw^\top \bfK \bfw
\end{eqnarray}
which yields the nice and desirable closed-form expression 
\begin{eqnarray}
\widehat{\bfw} = (\bfK+n\lambda\bfI_n)^{-1}\bfY.
\end{eqnarray}
The natural question that arises upon seeing this regularization framework is how to port or rather adapt the network data into it. As remarked by \cite{li2019}, the model inherent in \eqref{eq:regularized:HK:2} uses an intercept $\alpha$ that remains the same for all the observations in the training set $\sD$. Providentially, it turns out that making the intercept observation-dependent naturally provides a mechanism (doorway) for incorporating the network data into the model via some graph theoretical derivations. 
Such clever incorporation appear also elsewhere, namely \cite{Shalizi:2011},   \cite{James:2015}, \cite{li2019}\cite{Meghanathan:2019}, \cite{graph:2019:2}, \cite{Marques:graph:2017}, \cite{Shen:2017:Graph}. In our context, the use of network data translates into changing the model of \eqref{eq:basic:gaussian:regr:1} into the newer and more suitable model 

\begin{eqnarray}
\label{eq:basic:gaussian:regr:2}
p(y_i | f, \vx_i, \alpha, \bfw, \sigma^2) = \phi\left(y_i; \alpha_i+ \sum_{j=1}^{n}{\rw_j\krnl(\vx_i, \vx_j)}, \sigma^2\right) = \frac{1}{\sqrt{2\pi\sigma^2}}e^{-\frac{1}{2}\frac{\left(y_i-\alpha_i+ \displaystyle \sum_{j=1}^{n}{\rw_j \krnl(\vx_i, \vx_j)}\right)^2}{\sigma^2}},
\end{eqnarray}
where $\alpha_i$'s are then brought in as contributors to the model from the network data. We therefore end up having  $2n$ quantities to estimate, namely $n$ from the vector $\bfalpha$ and $n$ from the vector $\bfw$.  Clearly, with only $n$ observations in the data, estimating $2n$ quantities is inherently illposed. Regularization ends up being the approach used to circumvent this difficulty, with assumptions like $A_{uv}=1$ and  $A_{uu}=0$ in the adjacency matrix as in \cite{li2019}. Specifically, the resulting regularized kernel regression model, in its raw (canonical)  form is given by 
\begin{eqnarray}
\label{eq:regularized:HK:3}
\widehat{R}_{\lambda,n}(\bfalpha,\bfw) = (\bfY-\bfalpha -\bfK \bfw)^\top(\bfY-\bfalpha - \bfK \bfw) + \lambda \bfalpha^\top \bfL \bfalpha,
\end{eqnarray}
where the Gram matrix $\bfK$ of similarities is given by
\begin{eqnarray}
\label{eq:gram:K:1} \bfK=
\begin{bmatrix}
\krnl(\vx_1,\vx_1) & \krnl(\vx_1,\vx_2) & ... & \krnl(\vx_1,\vx_n)  \\
\krnl(\vx_2,\vx_1) & \krnl(\vx_2,\vx_2) & ... & \krnl(\vx_2,\vx_n)  \\
\krnl(\vx_3,\vx_1) & \krnl(\vx_3,\vx_2) & ... & \krnl(\vx_3,\vx_n)  \\
... & ... & ... & ...\\
\krnl(\vx_n,\vx_1) & \krnl(\vx_n,\vx_2) & ... & \krnl(\vx_n,\vx_n)  
\end{bmatrix}
\end{eqnarray}
the vector $n$-dimensional vector $\bfalpha=(\alpha_1,\alpha_2,...,\alpha_n)\in \mathbb{R}^n$ is the vector of the individual effects of every node,  the vector $\bfw=(\rw_1,\rw_2,...,\rw_n)^\top\in \mathbb{R}^n$ is the vector of regression coefficients, and 
\begin{eqnarray}
\label{eq:graph:reg:1}
    \lambda \bfalpha^\top \bfL\bfalpha=\sum_{(u,v)\in \bfE}(\alpha_u-\alpha_v)^2.
\end{eqnarray}
In the above, it is assumed that there a graph $\bfG=(\bfV,\bfE)$ captures the links between the observations from the network data, where $\bfV=\{1,2,...,n\}$ is the node set of the graph, and $\bfE\in \bfV \times \bfV$ is the edge set. The corresponding   adjacency matrix $\bfA\in \mathbb{R}^{n \times  n}$ is such that so the Laplacian of $\bfG$ is given by $ \bfL=\bfD-\bfA$, where $\bfD = diag(d_1,d_2,...,d_n)$ is the degree matrix. \cite{Vedanayak:M:2014}.  Straightforward computations of  derivatives with matrix differentiation \cite{Barnes:2006} wherever needed help obtain the estimates of both $\bfalpha$ and $\bfw$. It is straightforward to establish that the estimators are given by
\begin{eqnarray}
\label{eq:sol:1}
    (\widehat{\bfalpha}, \widehat{\bfw})=(\tilde{\bfK}^\top \tilde{\bfK} + \lambda \bfM)^{-1} \tilde{\bfK}^\top \bfY,
\end{eqnarray}
where $\bfY=(y_1,\cdots,y_n)$ is the $n$-dimensional vector of response values, $\tilde{\bfK}=[I_{n \times n},\bfK]$ and $\bfM=\left[\begin{array}{cc}
     L & 0_{n\times n}  \\
    0_{n\times p} & 0_{n\times n}
\end{array}\right].$
While the above straightforward adaptation of \cite{li2019} to kernel regression does indeed provide several advantages, we deemed it even better to combine both the traditional regularization on $\bfw$ with the on $\bfalpha$ coming from the network. As a result a more complete and certainly more powerful framework is derived as seen in the following regularized objective function
\begin{eqnarray}
\label{eq:regularized:HK:4}
\widehat{R}_{\lambda,n}(\bfalpha,\bfw) = (\bfY-\bfalpha -\bfK\bfw)^\top(\bfY-\bfalpha - \bfK\bfw) + \lambda\bfalpha^\top \bfL \bfalpha + \psi \bfw^\top \bfK \bfw.
\end{eqnarray}
Fortunately, the derivation of the corresponding is straightforward and yields
\begin{align}
    (\widehat \bfalpha, \widehat \bfw)=(\tilde{\bfK}^\top \tilde{\bfK} + \psi \bfN+\lambda\bfM)^{-1} \tilde{\bfK}^\top \bfY,
    \end{align}
where $\tilde{\bfK}=[I_{n \times n},\bfK]$, $\bfN=\left[\begin{array}{cc}
     0_{n\times n} & 0_{n\times n}  \\
    0_{n\times n} & I_{n\times n}
\end{array}\right]$ and $\bfM=\left[\begin{array}{cc}
    L & 0_{n\times n}  \\
    0_{n\times n} & 0_{n\times n}
\end{array}\right].$
It is remarkable how straightforward the adaptation of network cohesion to kernel regression turns out to be, especially considering how substantial the gain in predictive performance ends up being in practice as seen later in the computational demonstrations. As stated earlier, one of the greatest benefits of kernel methods remains the ability they provide to handle non numeric input spaces for which appropriate kernels (measures of similarity) are well defined.

\subsection{Obtaining Predictions in Training and Testing}
Since improvement in predictive performance is one of the main motivations for resorting to kernel methods, we now show just how straightforward also the prediction function is in this case. Having obtained $\widehat{\bfalpha}$ and $\widehat{\bfw}$, it is straightforward that for all $(\vx_i,y_i) \in \sD$
\begin{eqnarray}
\widehat{f}(\vx_i) = \widehat{\alpha}_i + \sum_{j=1}^n{\widehat{\rw}_j\krnl(\vx_i,\vx_j)}
\end{eqnarray}
so that for the whole training set $\sD$, the $n$-dimensional vector $\widehat{\bfY}$ of in-sample predicted (fitted) values is given by 
$$
\widehat{\bfY} = \tilde{\bfK}(\tilde{\bfK}^\top\tilde{\bfK}+\psi\bfN+\lambda \bfM)^{-1} \tilde{\bfK} \bfY = \bfH (\lambda)\bfY,
$$
where $\bfH(\lambda)=\tilde{\bfK}(\tilde{\bfK}^\top\tilde{\bfK}+\psi\bfN+\lambda \bfM)^{-1} \tilde{\bfK}$ is the so-called "hat" matrix.

Now, since one of the chief motivation supporting the adoption of kernel regression is the construction of optimal predictive regression models, it is important to focus a bit on the out-of-sample prediction mechanism. More specifically,
we seek to make $\mathbb{E}[\widehat{f}(\vx_{{\tt new}})]$ small, where
$$
\widehat{f}(\vx_{{\tt new}}) = \widehat{\alpha}_{{\tt new}} + \sum_{i=1}^n{\widehat{\rw}_{i} \bfK(\vx_{{\tt new}}, \vx_i)}
$$
With $n$ training observations and $m$ testing observations, the resulting Laplacian is $(n+m)\times (n+m)$ matrix that admits the decomposition of the form: $$\tilde{\bfL}=\left(\begin{array}{cc}
      \bfL_{ss} & \bfL_{st}  \\
      \bfL_{ts} & \bfL_{tt}
\end{array}\right)$$
where $\bfL_{ss}$ corresponds to the $n$ training observations, $\bfL_{tt}$  to the $m$ observations from the test set, and  $\bfL_{tt}$ corresponds the relationships between the members of the training set and those of the test set. The individual effect vector is $(\bfalpha_{s},\bfalpha_{t})$, where ${\bfalpha_{s}}$ is the estimator for $\bfalpha_1$ based on the training data. and our goal is predicting $\bfalpha_2$ denoted as $\hat{\bfalpha_2}$, what we do here is by minimizing the overall cohesion penalty, which is:
$$\bfalpha_2=\underset{\bfalpha_{t}}{{\tt argmin}}\Big\{(\widehat{\bfalpha_{s}},\bfalpha_{t})^\top \bfL (\widehat{\bfalpha}_{s},\bfalpha_{t})\Big\}
$$
This gives us:
$$
\widehat{\bfalpha}_{t}=-\bfL_{tt}^{-1}\bfL_{ts}\widehat{\bfalpha}_{s}
$$
Given the estimate $\widehat{\bfalpha}_{t}$, it is straightforward to make out of sample predictions.

\section{Computational Explorations and Demonstrations}
\begin{figure}[!ht]
    \centering
    \begin{subfigure}[b]{0.49\textwidth}
        \includegraphics[scale=.5,width=\textwidth]{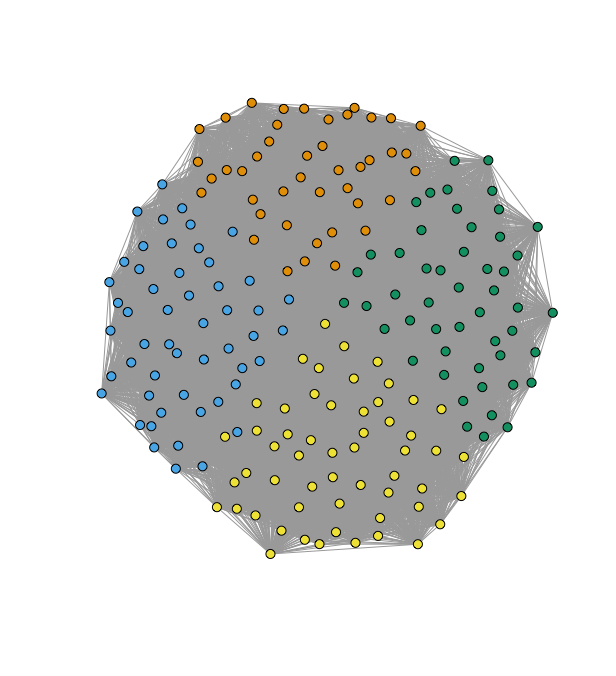}
        \caption{uniform network}
        \label{fig:uniformplot}
    \end{subfigure}
    \begin{subfigure}[b]{0.49\textwidth}
        \includegraphics[scale=.5,width=\textwidth]{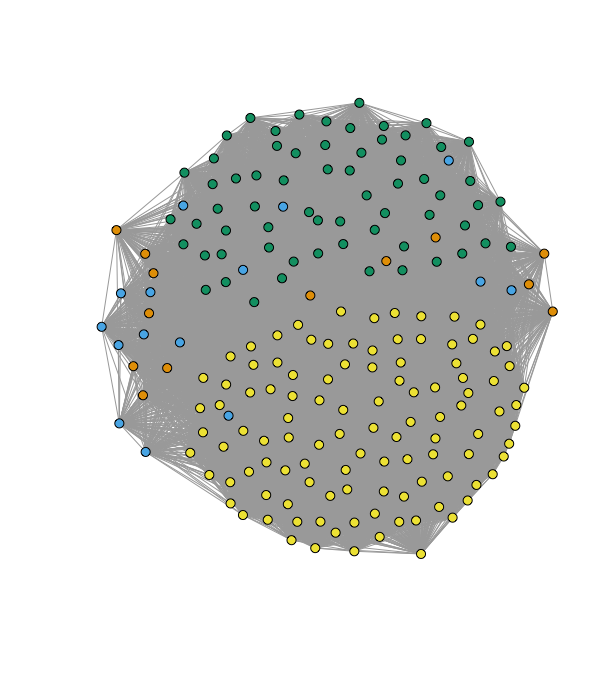}
        \caption{tight network}
        \label{fig:tightplot}
    \end{subfigure}

        \begin{subfigure}[b]{0.49\textwidth}
        \includegraphics[scale=.5,width=\textwidth]{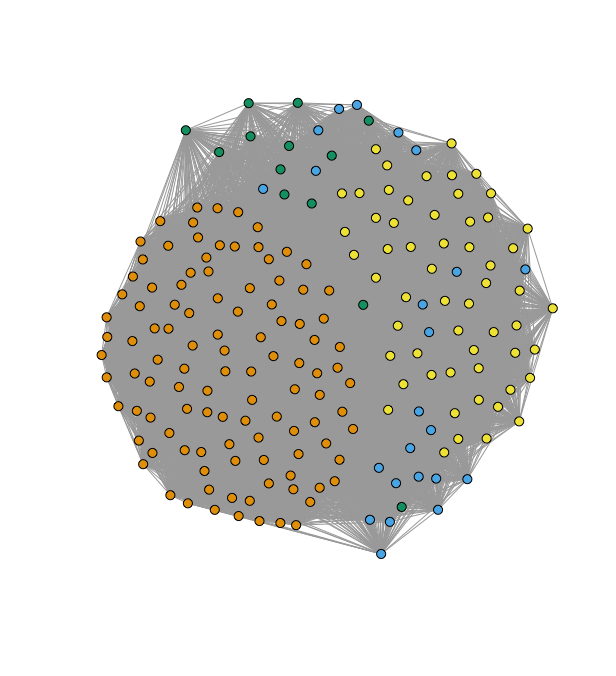}
        \caption{open network}
        \label{fig:openplot}
    \end{subfigure}
    \begin{subfigure}[b]{0.49\textwidth}
        \includegraphics[scale=.5,width=\textwidth]{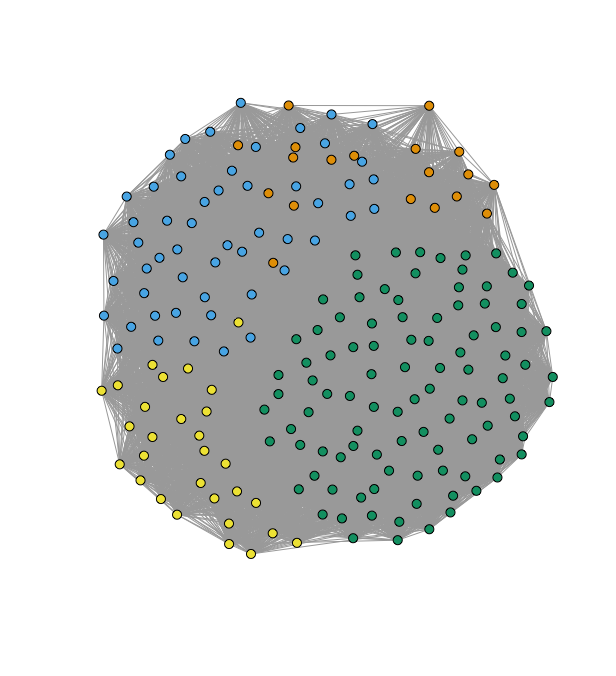}
        \caption{wide open network}
        \label{fig:wideopenplot}
    \end{subfigure}
    \caption{4 types of networks}
      \label{fig:types of networks}
\end{figure}

In this section, we used simulated data and real world data to test the performance of the model, we compares kernel regression with network cohesion model with linear regression with network cohesion model, multilinear regression model, support vector machine, relevance vector machine, and Gaussian process machine. 
The simulate data contains 200 observations and each sample contains 2 predictors and 1 respond variable, are generated from a multivariate random normal distribution, are done by R function "rnorm", also generate the error add in to the data. In order to test variety networks, we generate 4 different kinds of network types so-called uniform, tight, open and wide-open, as shown in the figure\ref{fig:types of networks}. Each network is generated from a sample size of 200 and into 4 groups. Uniform network is generated from a uniform distribution, tight, open and wide-open networks are generated from dirichlet distributions with the help of R package "MCMCpack" \cite{MCMC:R}. 

\subsection{Performance of different machines}
In this section we compare the 10 machines fitted to the simulated data with 4 different kinds of networks. In Table \ref{table:training error} and Figure \ref{table:training error table} show the comparison of the training error, which contains 100 observations, based on the tables and figures, we could have a clear results, the machines considered about the network cohesion are better than those traditional machines, and since there are some nonlinearity in the data, the linear models are not comparable as expected. Amount all the machines, kernel model with tangent kernel is the winner of all the machines and also the best even consider different kinds of networks.

\begin{table}
\caption{Training error. LIN: linear regression with network cohesion, COS: kernel regression with network cohesion models that use cosine kernel, RBF: Gaussian kernel, LPC: laplace kernel, NN: hyperbolic tangent kernel, POL: polynomial kernel, MLR: multilinear regression, SVM: support vector machine, RVM: relevance support machine and GP: Gaussian processes for regression}\label{table:training error table}
\begin{tabular}{lllllllllll}
\toprule
    SN  & MLR & LIN & COS & RBF & LPC & NN & POL  & SVM & RVM   & GP  \\ 
\midrule
Uf & 1.70 &  1.10  & 0.27 &  0.19 &   0.20 & \textbf{0.17} &  0.27 & 1.77 & 0.92 & 1.80 \\ 
Ti & 1.67 &  1.14   &  0.31 &  0.22 &  0.23 & \textbf{0.19} &    0.31 & 1.72 & 0.90 & 1.76 \\
Wo & 1.66  & 1.19 &   0.35 &  0.25 &  0.26 & \textbf{0.22} &   0.35 & 1.71 & 0.90 & 1.75 \\
Op & 1.65 &  1.09  & 0.27 &  0.20 &  0.21 & \textbf{0.18} &  0.27 & 1.70 & 0.93 & 1.74 \\
\bottomrule
\end{tabular}
\end{table}

\begin{figure}[!ht]
    \centering
    \begin{subfigure}[b]{0.49\textwidth}
        \includegraphics[width=\textwidth]{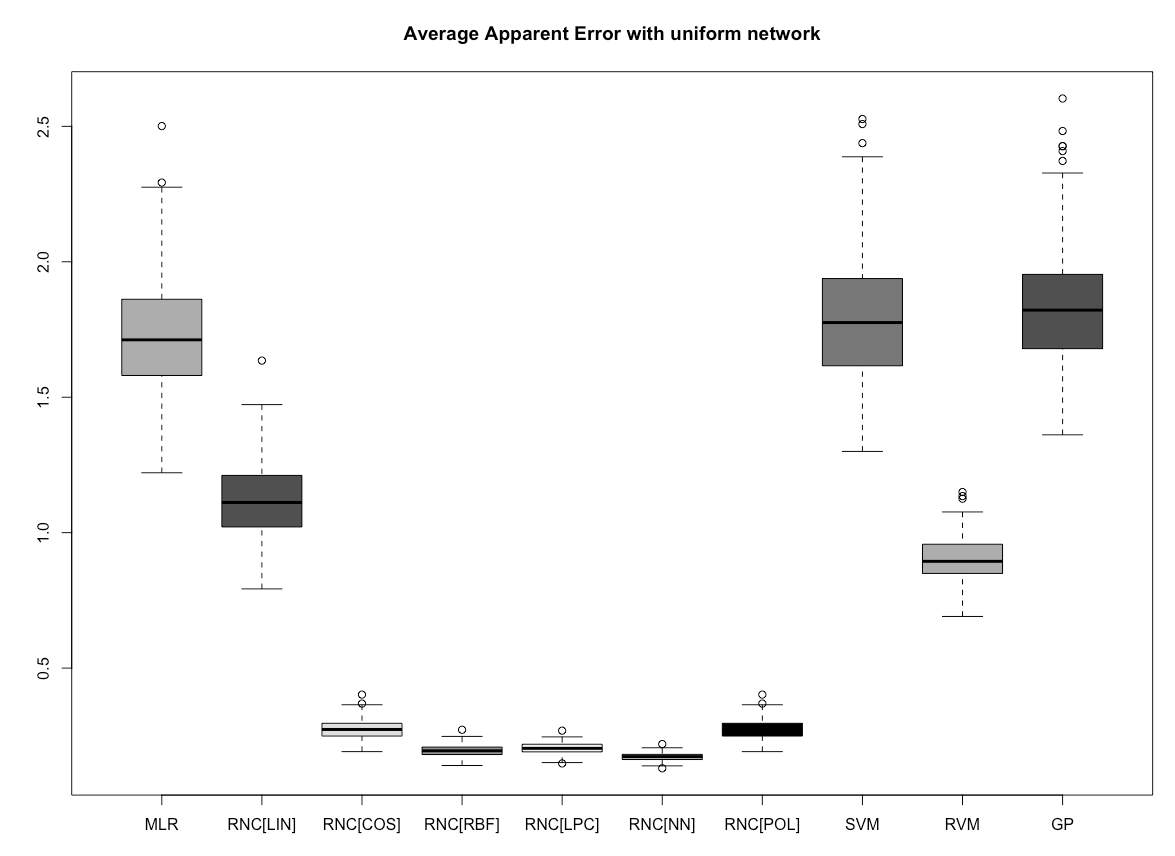}
        \caption{uniform}
        \label{fig:sigmapost}
    \end{subfigure}
    \begin{subfigure}[b]{0.49\textwidth}
        \includegraphics[width=\textwidth]{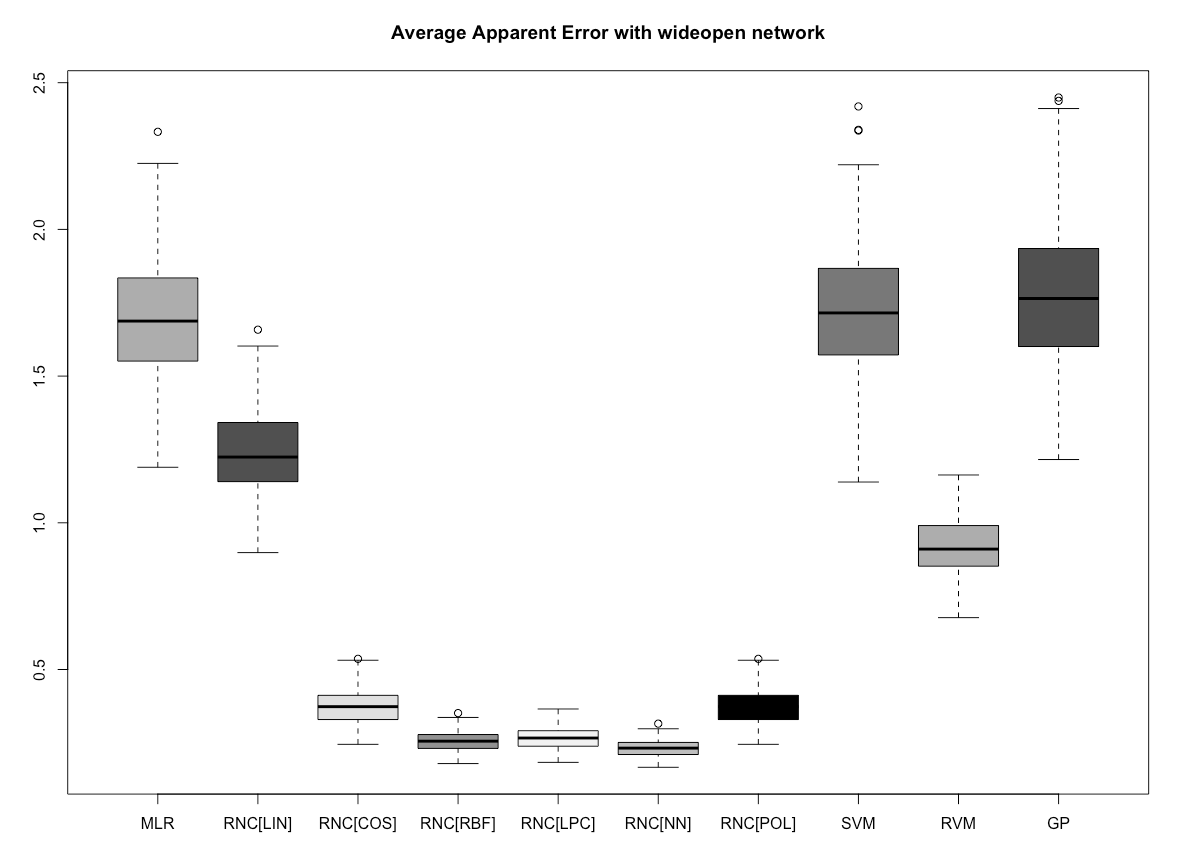}
        \caption{wideopen}
        \label{fig:comparepost}
    \end{subfigure}

        \begin{subfigure}[b]{0.49\textwidth}
        \includegraphics[width=\textwidth]{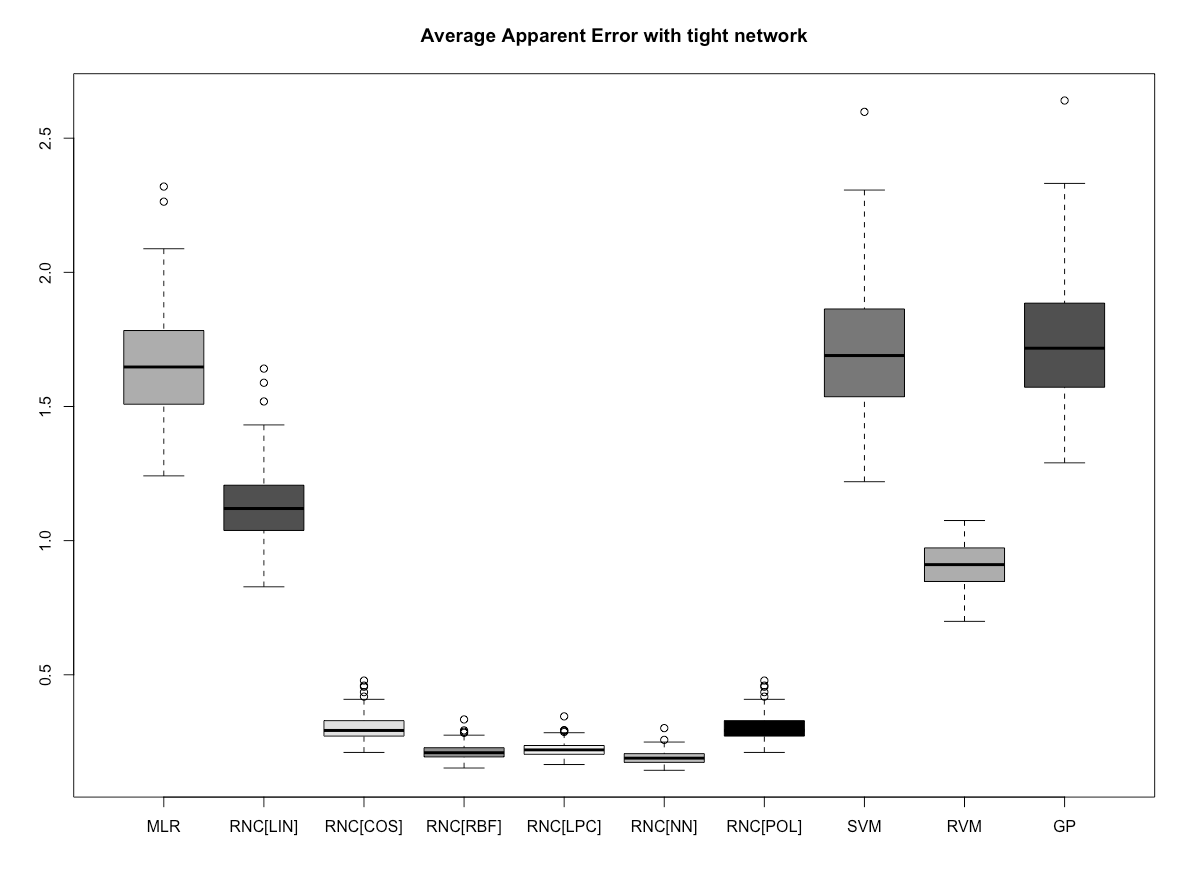}
        \caption{tight}
        \label{fig:sigmapost}
    \end{subfigure}
    \begin{subfigure}[b]{0.49\textwidth}
        \includegraphics[width=\textwidth]{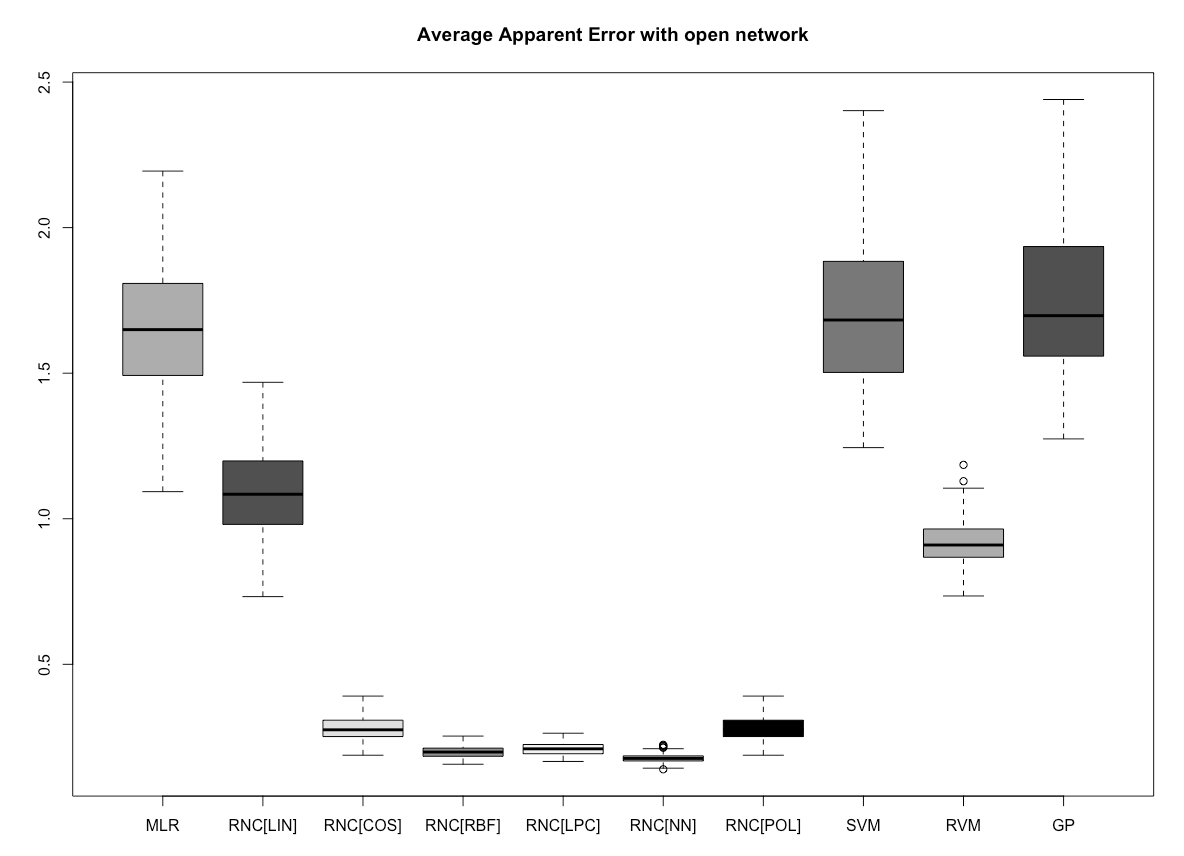}
        \caption{open}
        \label{fig:comparepost}
    \end{subfigure}
    \caption{the training error}
      \label{table:training error}
\end{figure}

In table \ref{table:test error} and Figure \ref{table:test error table}, which shown the performance of the machines on the test data, amount 10 machines and 4 kinds of networks the winner is still kernel machine with tangent kernel, but kernel machines with cosine and polynomial kernels are not as good as they are in the training set performance. some traditional methods start to doing better, like support vector machine gets really comparable to the kernel machines, but still not as good as some kernel machines. So based on the test error, we can see if we choose the right kernel for the kernel machines, we could improve the prediction performance when considering the network cohesion.

\begin{table}
\caption{Test error. LIN: linear regression with network cohesion, COS: kernel regression with network cohesion models that use cosine kernel, RBF: Gaussian kernel, LPC: laplace kernel, NN: hyperbolic tangent kernel, POL: polynomial kernel, MLR: multilinear regression, SVM: support vector machine, RVM: relevance support machine and GP: Gaussian processes for regression}\label{table:test error table}
\begin{tabular}{lllllllllll}
\toprule
    SN  & MLR & LIN & COS & RBF & LPC & NN & POL  & SVM & RVM   & GP  \\ 
\midrule
Uf & 1.77   &1.77  & 1.78  & 1.34  & 1.38 & \textbf{1.26}  & 1.77 & 1.36 & 1.43 & 1.41 \\ 
Ti & 1.75  & 1.75  & 1.76  & 1.31  & 1.36 & \textbf{1.23}  & 1.75 & 1.34 & 1.39 & 1.38 \\
Wo & 1.75  & 1.75 &  1.76 &  1.29 &  1.34 &  \textbf{1.22} &  1.76  & 1.32 & 1.36  & 1.36 \\
Op & 1.78 &  1.77  & 1.79  & 1.37 &  1.42 & \textbf{1.28}  & 1.78 & 1.42 & 1.46 & 1.45 \\
\bottomrule
\end{tabular}
\end{table}

\begin{figure}[!ht]
     \centering
    \begin{subfigure}[b]{0.49\textwidth}
        \includegraphics[width=\textwidth]{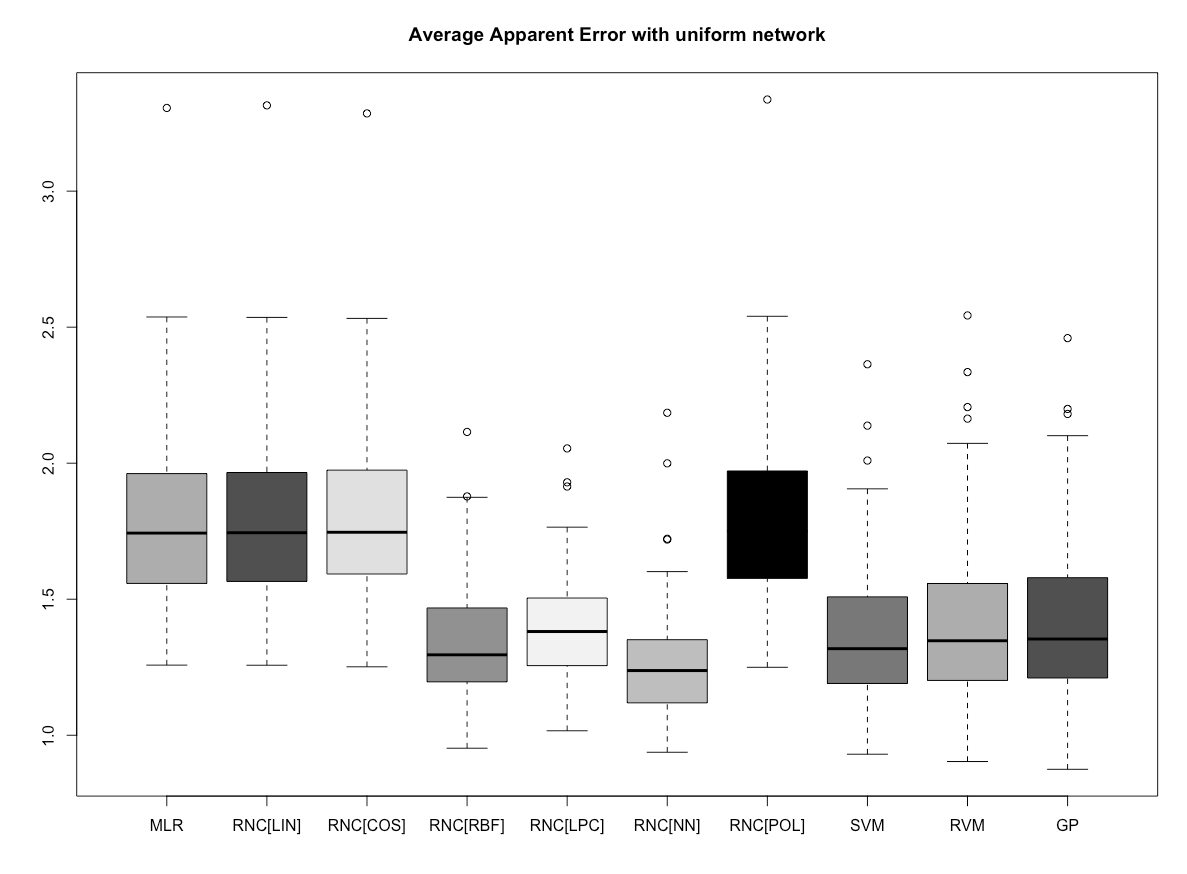}
        \caption{uniform}
        \label{fig:sigmapost}
    \end{subfigure}
    \begin{subfigure}[b]{0.49\textwidth}
        \includegraphics[width=\textwidth]{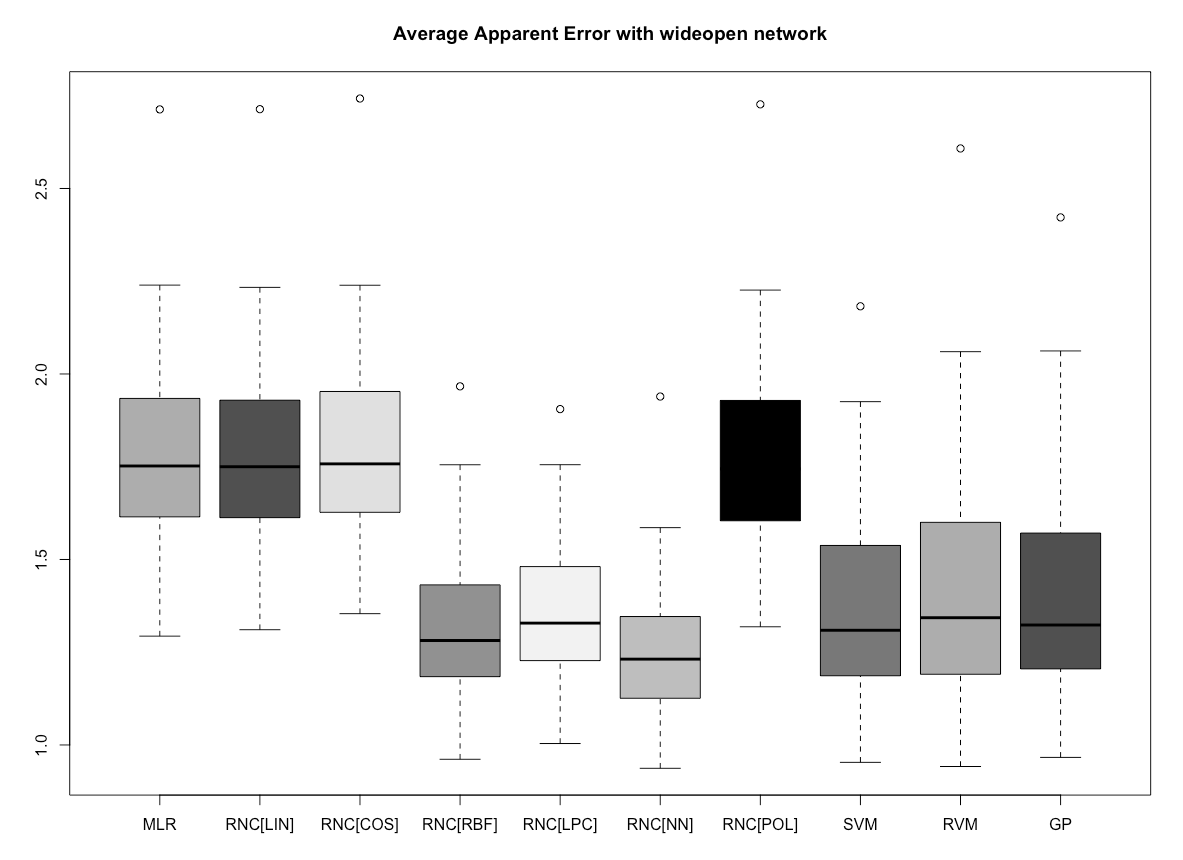}
        \caption{wideopen}
        \label{fig:comparepost}
    \end{subfigure}

        \begin{subfigure}[b]{0.49\textwidth}
        \includegraphics[width=\textwidth]{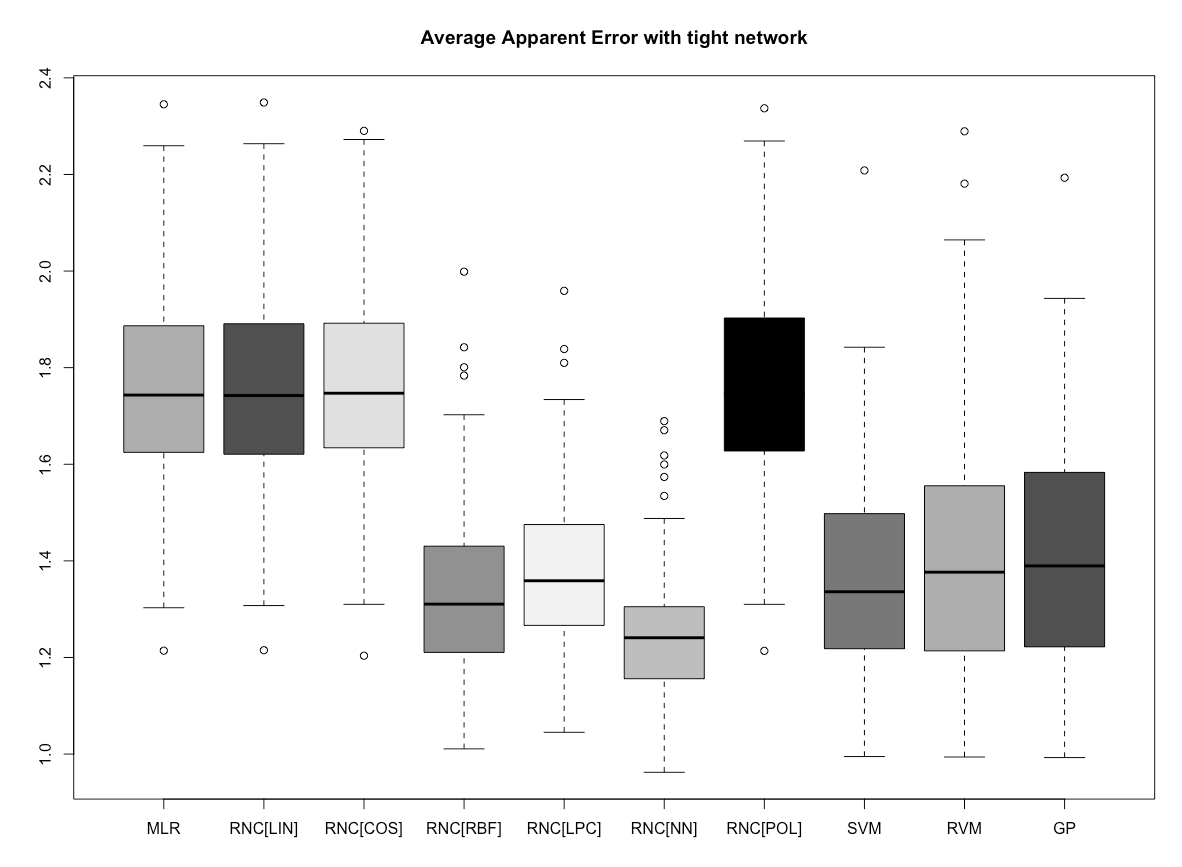}
        \caption{tight}
        \label{fig:sigmapost}
    \end{subfigure}
    \begin{subfigure}[b]{0.49\textwidth}
        \includegraphics[width=\textwidth]{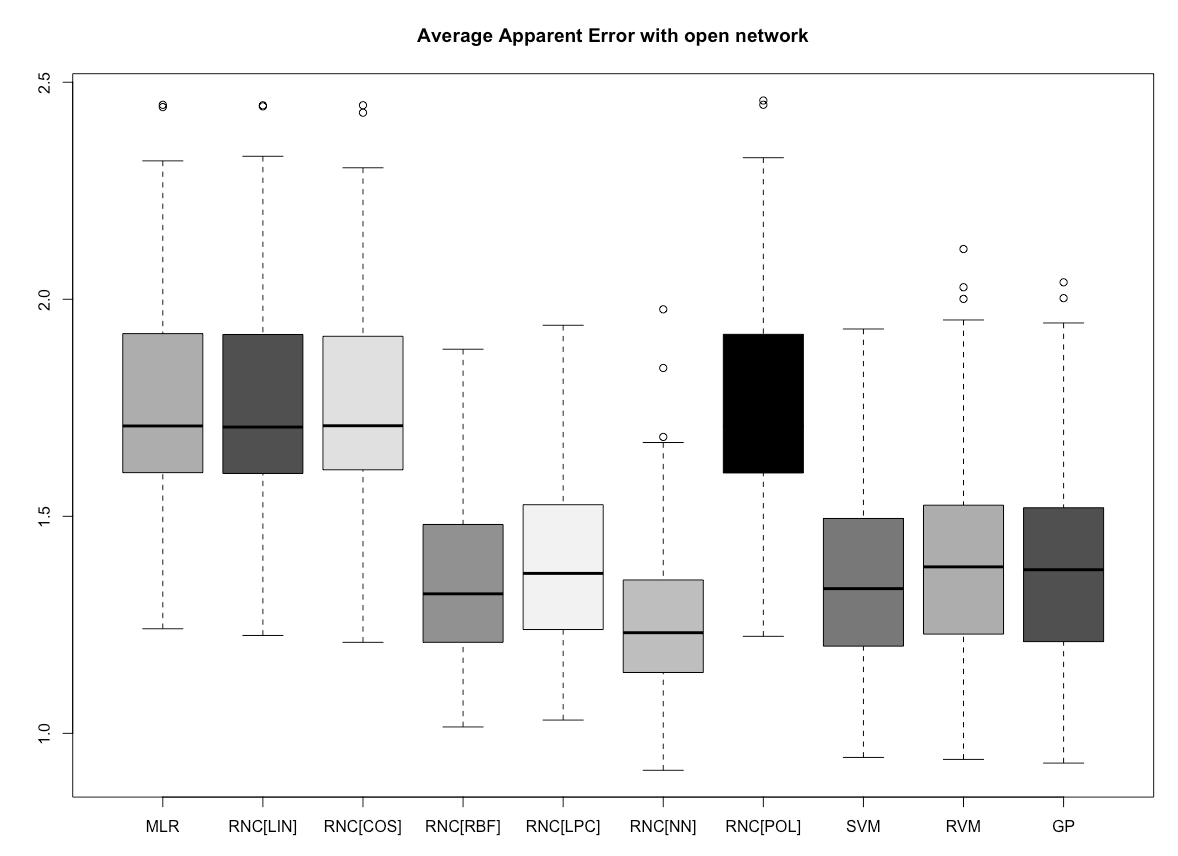}
        \caption{open}
        \label{fig:comparepost}
    \end{subfigure}
    \caption{the training error}
      \label{table:test error}
\end{figure}

\subsection{Applications}
In this section, we imply our method to the 'Teenage Friends and Lifestyle Study' data \cite{Bush:1997:data}. The social network data were collected in the Teenage Friends and Lifestyle Study. Friendship network data and substance use were recorded for a cohort of pupils in a school in the West of Scotland. The panel data were recorded over a three year period starting in 1995, when the pupils were aged 13, and ending in 1997. A total of 160 pupils took part in the study, 129 of whom were present at all three measurement points. The friendship networks were formed by allowing the pupils to name up to twelve best friends. Pupils were also asked about substance use and adolescent behavior associated with, for instance, lifestyle, sporting behavior and tobacco, alcohol and cannabis consumption. The school was representative of others in the region in terms of social class composition. 

In this application task, we are focus on "how often does one using Alcohol?", among all the factors, we choose "age", "sex", "romantic", "family smoking", "money"(their allowance), "sport", "church"  8 factors as our variables. The data is divided into training and test set, which is 70 \% and 30 \%. We compare the MSE.
\begin{figure}[!ht]
	\centering
		\includegraphics[scale=.5]{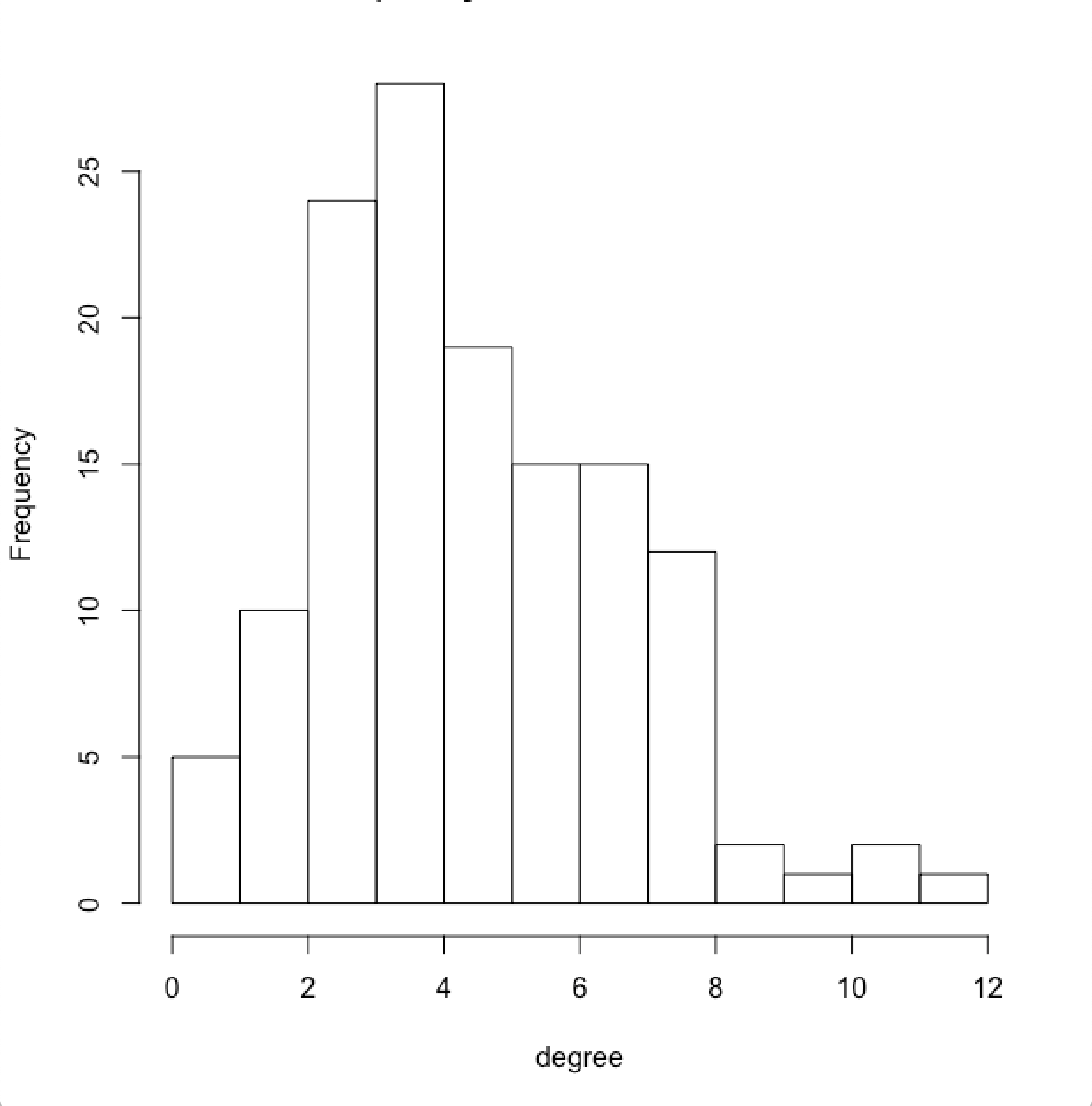}
	\caption{the histogram of the numbers of friends, most students have 3-5 friends in school}
	\label{fig:frequency}
\end{figure}

\begin{figure}[!ht]
	\centering
	\includegraphics[scale=.5]{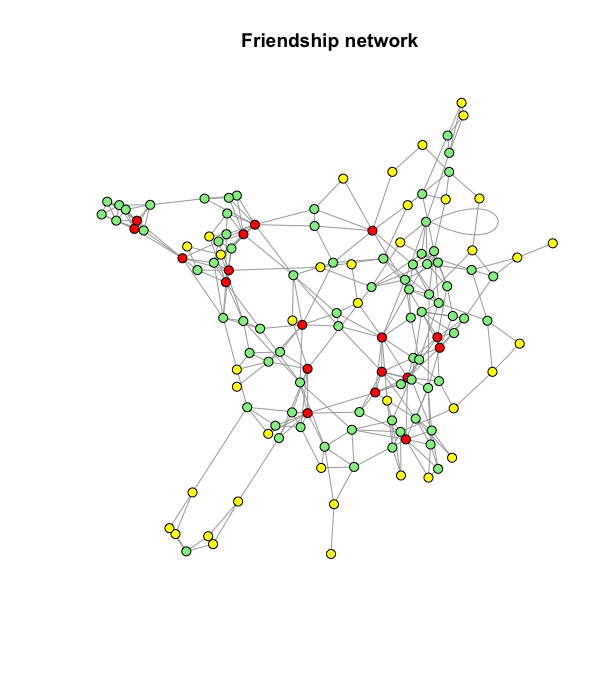}
	\caption{the network of all the students, there are 3 colors in the plots, yellow means one has less than 4 friends, green mean one has 4-8 friends, red means one has more than 8 friends.}
    \label{fig:network}
\end{figure}

Based on the table \ref{table:training error}, we could see all the network cohesion models are doing better than the rest without consider about the network. From the table \ref{table:test error}, we could see the RNC-linear model is slightly better than MLR, and also they all behave better than SVM, RVM and GP. RNC-kernel models mostly have the small MSE, except with the polynomial kernel, among all the machines, RNC with Gaussian RBF kernel is the best machine. SO the result shows the models with network cohesion could lead to some improvement, especially the kernel machines, when we choose the right kernel, we could definitely get a improved results.

\begin{table}[!ht]
\caption{Training error. LIN: linear regression with network cohesion, COS: kernel regression with network cohesion models that use cosine kernel, RBF: Gaussian kernel, LPC: laplace kernel, NN: hyperbolic tangent kernel, POL: polynomial kernel, MLR: multilinear regression, SVM: support vector machine, RVM: relevance support machine and GP: Gaussian processes for regression}\label{table:training error}
\begin{tabular}{lllllllllll}
\toprule
    SN  & MLR & LIN & COS & RBF & LPC & NN & POL  & SVM & RVM   & GP  \\ 
\midrule
Training Error & 0.5617 &  0.0679 &  0.0074 &  0.0071 &  0.0074 & 0.0075 &  0.0081 & 1.6559 & 0.3772 & 1.3399 \\ 
\bottomrule
\end{tabular}
\end{table}

\begin{table}[!ht]
\caption{Test error. LIN: linear regression with network cohesion, COS: kernel regression with network cohesion models that use cosine kernel, RBF: Gaussian kernel, LPC: laplace kernel, NN: hyperbolic tangent kernel, POL: polynomial kernel, MLR: multilinear regression, SVM: support vector machine, RVM: relevance support machine and GP: Gaussian processes for regression}\label{table:test error}
\begin{tabular}{lllllllllll}
\toprule
    SN  & MLR & LIN & COS & RBF & LPC & NN & POL  & SVM & RVM   & GP  \\ 
\midrule
test error & 0.5809  & 0.5796 &  0.5601  & 0.5148 &  0.5262 &  0.579 &  0.6688 & 0.5949 & 0.6066 & 0.6038 \\ 
\bottomrule
\end{tabular}
\end{table}

\begin{figure}[!ht]
    \centering
    \begin{subfigure}[b]{0.49\textwidth}
        \includegraphics[width=\textwidth]{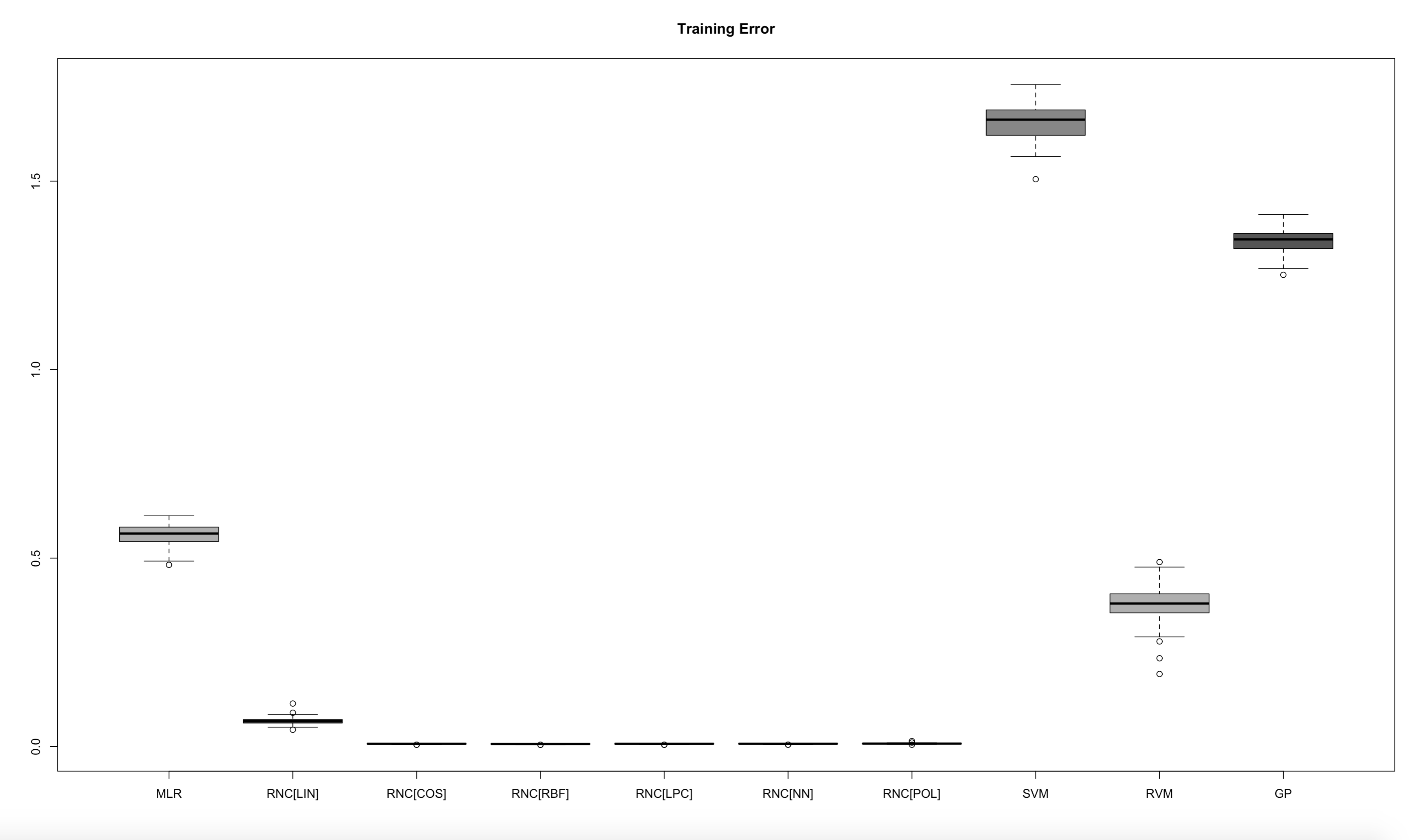}
        \caption{training error }
        \label{fig:sigmapost}
    \end{subfigure}
    \begin{subfigure}[b]{0.49\textwidth}
        \includegraphics[width=\textwidth]{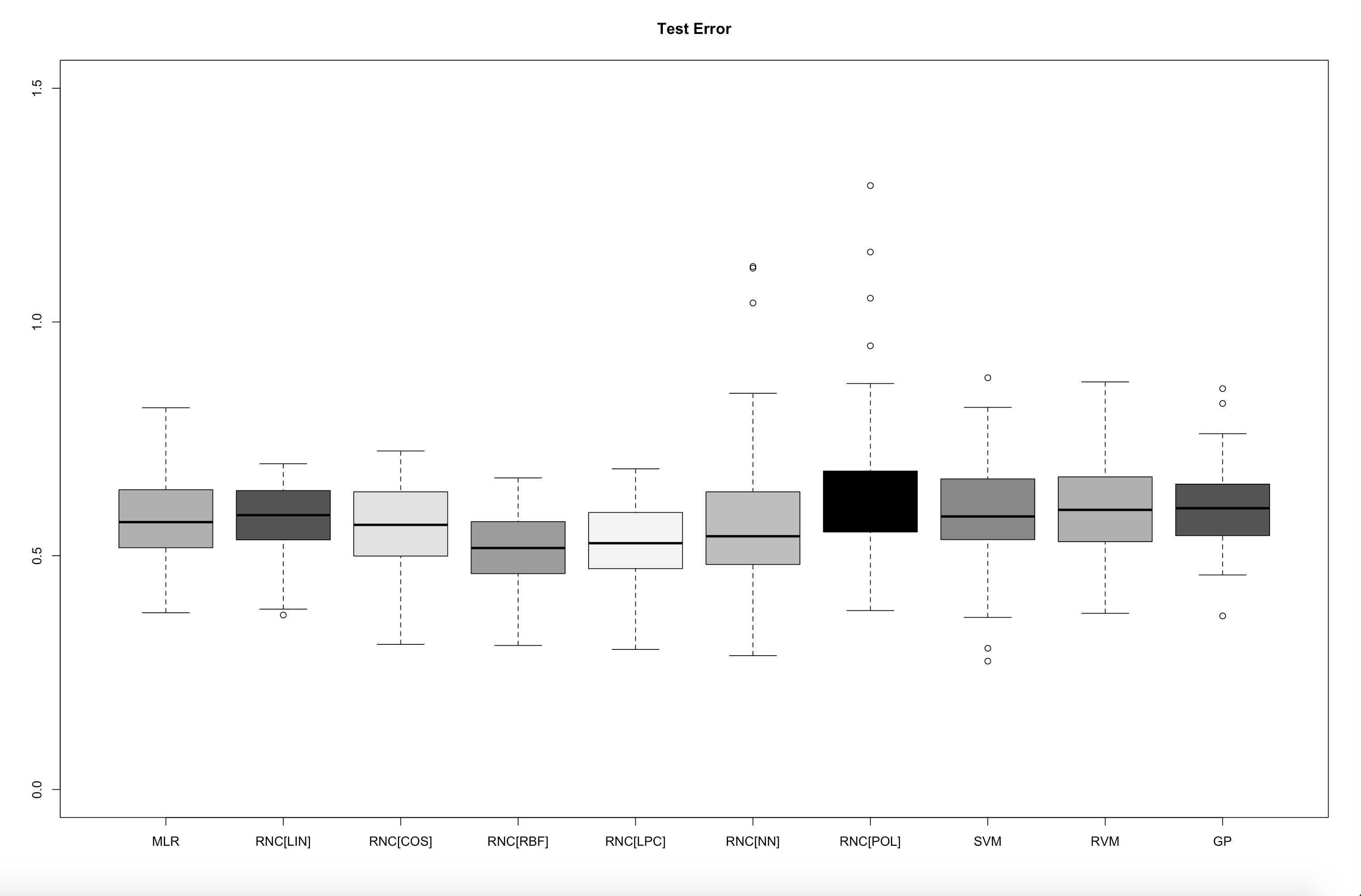}
        \caption{test error }
        \label{fig:comparepost}
    \end{subfigure}
    \caption{The left image shows the mean MSE for the training set in the 45 runs.
      The right image shows the mean MSE for the test set in the 45 runs}
      \label{fig:simulationregression}
\end{figure}

\section{Conclusion}
This paper has proposed and developed the incorporation of graph theoretic constraints into kernel regression with the finality of improving the predictive performance with information from network data. We demonstrated throughout that this appealing adaptation does yield models with better predictive performances and greater modeling flexibility. Thanks to the straightforward nature of the derived model, we anticipate that our work has the potential of being widely used on a wide variety of applications. As part of our future research on this fascinating theme, we intend to explore more types of graphs and network topologies, but also crucially, we plan on providing a thorough theoretical justification of the performances obtained.

\bibliographystyle{unsrt}  
%\bibliography{references}  %%% Remove comment to use the external .bib file (using bibtex).
%%% and comment out the ``thebibliography'' section.
%%% Comment out this section when you \bibliography{references} is enabled.
\newpage
\bibliography{references}

\newpage
\appendix
\section{My Appendix}
{\sf Proof}: By using the matrix derivative, 
\begin{align*}
    & \frac{d L(\bfalpha,\bfw)}{d \bfalpha} =-2 \bfK^\top (\bfY-\bfK\bfw-\bfalpha)+2 \psi\bfI_{n\times n} \bfw=0 \\
    & \frac{d L(\bfalpha,\bfw)}{d \bfw} =-2\bfI_{n\times n}\bfK^\top(\bfY-\bfK\bfw-\bfalpha)+2\lambda \bfL \bfalpha=0   
\end{align*}

Solve the above equations:
\begin{align*}
    & \bfK^\top \bfY-\bfK^\top\bfK \bfw-\bfK^\top \bfalpha-\psi \bfI \bfw=0 \\
    & \bfI \bfY-\bfK \bfw- \bfI \bfalpha- \lambda \bfL \bfalpha=0
\end{align*}

which can be written as:
\begin{align*}
    & \bfK^\top\bfY=(\bfK^\top\bfK +\psi \bfI)\bfw+\bfK^\top \bfalpha \\
    & I \bfY=\bfK \bfw+ (\bfI + \lambda \bfL )\bfalpha
\end{align*}

The result could be written as matrix form as:
$$\left(\begin{array}{cc}
     I_n \\
    \bfK^\top 
\end{array}\right)Y=\left(\begin{array}{cc}
     \bfI_n + \lambda \bfL  & \bfK  \\
    \bfK^\top & \bfK^\top\bfK+\psi \bfI_n
\end{array}\right)\left(\begin{array}{cc}
     \bfalpha \\
    \bfw
\end{array}\right)$$

which equal to 
$$\left(\begin{array}{cc}
     I_n \\
    \bfK^\top 
\end{array}\right)\bfY=\left[\left(\begin{array}{cc}
     \bfI_n & \bfK  \\
    \bfK & \bfK^\top\bfK
\end{array}\right)+
\lambda\left(\begin{array}{cc}
      \bfL & \nmathbf{0}  \\
    \nmathbf{0} & \nmathbf{0}
\end{array}\right)+
\psi\left(\begin{array}{cc}
      \nmathbf{0} & \nmathbf{0}  \\
    \nmathbf{0} & \bfI_n
\end{array}\right)\right]
\left(\begin{array}{cc}
     \bfalpha \\
    \bfw
\end{array}\right)$$

\end{document}